\documentclass[10pt,twocolumn,letterpaper]{article}

\usepackage{iccv}
\usepackage{times}
\usepackage{epsfig}
\usepackage{graphicx}
\usepackage{cuted}
\usepackage{capt-of}
\usepackage{caption}
\usepackage{subcaption}
\usepackage{amsmath}
\usepackage{amssymb}
\usepackage{bm}
\usepackage{bbm}
\usepackage{mathtools}
\usepackage{booktabs} 
\usepackage{lipsum}
\usepackage{xcolor}
\usepackage{bbding}
\usepackage{colortbl}
\usepackage{enumitem}
\setenumerate[1]{itemsep=0pt,partopsep=0pt,parsep=\parskip,topsep=5pt}
\setitemize[1]{itemsep=0pt,partopsep=0pt,parsep=\parskip,topsep=5pt}
\setdescription{itemsep=0pt,partopsep=0pt,parsep=\parskip,topsep=5pt}

\DeclareMathOperator*{\argmin}{\arg\!\min}
\DeclarePairedDelimiterX{\js}[2]{}{}{%
  #1\;\delimsize\|\;#2%
}

\newcommand\blfootnote[1]{%
  \begingroup
  \renewcommand\thefootnote{}\footnote{#1}%
  \addtocounter{footnote}{-1}%
  \endgroup
}

\usepackage[pagebackref,breaklinks,colorlinks]{hyperref}

\iccvfinalcopy 

\def\iccvPaperID{****} 

\ificcvfinal\fi

\begin{document}

\title{Mojito: LLM-Aided Motion Instructor with Jitter-Reduced Inertial Tokens}

\author{
    Ziwei Shan\textsuperscript{1,*}\qquad
    Yaoyu He\textsuperscript{1,*}\qquad
    Chengfeng Zhao\textsuperscript{1,*,\textdagger}\qquad
    Jiashen Du\textsuperscript{1}\qquad
    Jingyan Zhang\textsuperscript{1}\\
    Qixuan Zhang\textsuperscript{1,2}\qquad
    Jingyi Yu\textsuperscript{1,\textdaggerdbl}\qquad
    Lan Xu\textsuperscript{1,\textdaggerdbl}\\
    \textsuperscript{1}ShanghaiTech University\qquad
    \textsuperscript{2}Deemos Technology\\
    {\tt\footnotesize \{shanzw2022,heyy2022,zhaochf2022,dujsh2022,zhangjy7,zhangqx1,yujingyi,xulan1\}@shanghaitech.edu.cn}
}

\maketitle
\ificcvfinal\thispagestyle{empty}\fi

\begin{strip}\centering
    \vspace{-45px}
    \captionsetup{type=figure}
    \includegraphics[width=\textwidth]{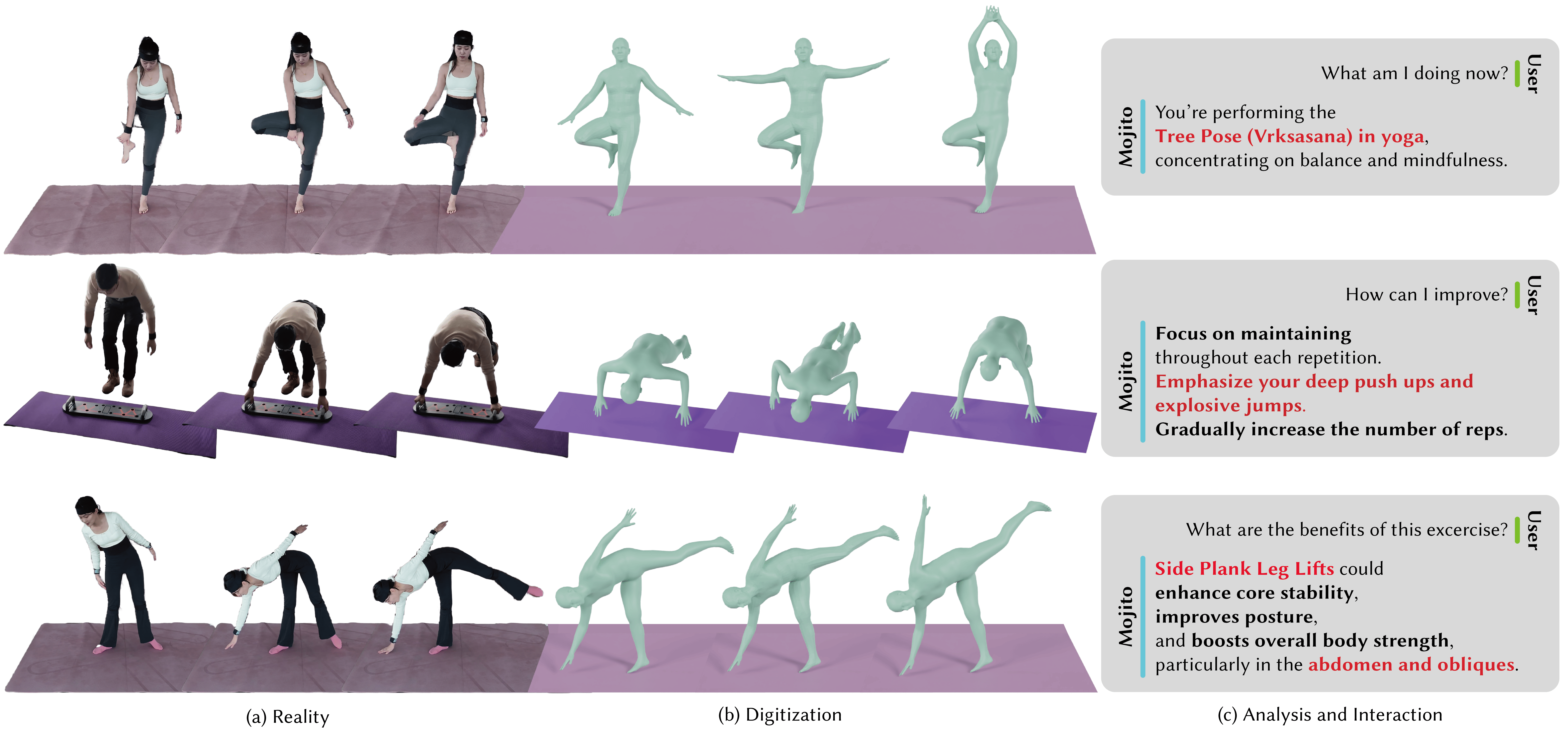}
    \vspace{-15px}
    \caption{Our \textit{Mojito} produces real-time human motion capture and online motion analysis, from six inertial measurement units (IMUs). (a) Human who performs exercise wearing IMU sensors in reality. (b) Digitalized human motion from six IMU sensor signals. (c) Motion recognition, analysis and instruction feedback.}
\label{fig:teaser}
\vspace{-10px}
\end{strip}

\blfootnote{\textsuperscript{*}Equal contributions}
\blfootnote{\textsuperscript{\textdagger}Project lead}
\blfootnote{\textsuperscript{\textdaggerdbl}Corresponding author}

\begin{abstract}
Human bodily movements convey critical insights into action intentions and cognitive processes, yet existing multimodal systems primarily focused on understanding human motion via language, vision, and audio, which struggle to capture the dynamic forces and torques inherent in 3D motion. Inertial measurement units (IMUs) present a promising alternative, offering lightweight, wearable, and privacy-conscious motion sensing. However, processing of streaming IMU data faces challenges such as instable wireless transmission, sensor noise, and drift, limiting their utility for long-term real-time motion capture (MoCap), and more importantly, online motion analysis. 
To address these challenges, we introduce Mojito, an intelligent motion agent that integrates inertial sensing with large language models (LLMs) for interactive motion capture and behavioral analysis. The core innovation of Mojito lies in a jitter-reduced inertial token representation with a novel IMU signal encoding framework, and an extended language model involving inertial tokens. By employing VQVAE, Mojito learns a discrete latent space of continuous IMU signals, mitigating sensor noise and drift through quantization. The inertial tokens are then aligned with inductive bias of natural language and mapped to textual semantics to enhance compatibility with LLMs, enabling efficient sequence modeling. 
To support domain-specific applications, Mojito further incorporates tunable LoRA adapters, facilitating personalized feedback tailored to roles such as fitness trainers or rehabilitation therapists. 
Extensive experiments demonstrate that Mojito outperforms existing IMU-based methods in motion capture under noisy conditions, and achieves comparable behavior analysis capability compared to large vision-language models. The user study further highlights its practical effectiveness in various scenarios as a versatile tool for intelligent human-agent interaction. Our code and data will be released at \href{https://koyui.github.io/mojito/}{our project page}.
\end{abstract}

\section{Introduction}
\label{sec:intro}

Human motion encapsulates rich information about action intentions and thought processes of us humans, serving as a crucial foundation for understanding human behavior patterns. Inertia, a contactless and continuously measurable physical quantity from IMUs, can directly reflect the dynamic forces and kinematic states underlying human movement. This measurement enables the reconstruction of physically consistent motion in virtual environments, transforming physical actions into analyzable digital representations. However, merely replicating motion in virtual spaces remains insufficient. Intelligent systems should also provide real-time feedback during human-computer interactions to enhance behavioral understanding and facilitate self-improvement. 
Therefore, a motion agent capable of real-time motion reconstruction and online behavior analysis becomes vital for various application scenarios such as exercising, rehabilitation, and skill development. Ideally, the capturing and analysis process should be user-friendly, highly accessible, and intuitive for interactions, just like modern conversational AI agents.

Recent advances in large language models (LLMs), have driven significant progress in multimodal intelligent systems, enabling natural language interaction across text, vision, audio, and even human motion expressed in SMPL~\cite{SMPL2015} parameters. However, they struggle to capture the dynamic forces and torques governing three-dimensional movement. Existing parametric motion representations, though useful for approximating body poses, omit critical temporal derivatives like velocity and acceleration, limiting their capacity to model the physics underlying motion. In contrast, IMUs overcome these limitations by providing wearable, high-frequency measurements of acceleration, angular velocity, and rotation, thereby offering spatio-temporally precise motion characterization. Therefore, for building an intelligent multimodal system that is capable of understanding and analyzing human motions, it is crucial to integrate and align the IMU modality with foundational perception modalities such as natural language. Nonetheless, naively taking SMPL~\cite{SMPL2015} as an intermediary to link raw IMU signals with natural language is suboptimal, since it inevitably discards certain raw signal patterns during parameterization. Moreover, achieving true multimodal alignment requires deeper integration of IMU data with foundational perception modalities, such as natural language investigated in this work.

In the field of computer graphics, IMU has become an essential tool for real-time human motion capture~\cite{noitom,Movella,mocopi} due to its practicality. Unlike camera-based systems for human mesh recovery, IMUs offer sparse, lightweight, occlusion-resistant sensing while preserving user privacy. Early IMU-based methods primarily relied on traditional optimization strategies to estimate human kinematic poses~\cite{von2017SIP}. More recent approaches, termed ``inertial posers"~\cite{huang2018DIP,TransPoseSIGGRAPH2021,PIPCVPR2022,van2024diffusionposer,yi2024pnp}, utilized data-driven neural networks to directly translate raw IMU signals into parametric body models~\cite{SMPL2015}, enabling wearable and efficient motion capture. However, existing methods remain limited to motion reconstruction, lacking higher-level analysis and contextual understanding of human movements. Advancing beyond basic capture capabilities, an intelligent motion agent capable of real-time feedback and multi-turn interaction with users could unlock transformative applications in healthcare, education, and digital fitness. For instance, text-based conversational interfaces could democratize access to skill development—novices in specialized exercises or rehabilitation routines might receive instant, tailored guidance through intuitive language interactions. Such a system would reduce learning barriers, lower costs, and enhance accessibility for diverse user groups.

In this work, we present Mojito, a novel IMU-based motion intelligence agent for real-time, continuous human motion capture and online analysis. Due to the inherent limitations of IMU sensors such as drift, cumulative errors, and external noise from connectivity or transmission issues, the feasibility of IMU-based systems in reliable and long-term motion analysis is hindered. To address these limitations, we introduce a noise-robust approach that diverges from prior methods. Specifically, instead of continuous representation, we encode IMU signals into a discrete latent space, within which the quantization strategy reduces jitter by mapping continuous IMU streams to fixed token sequences. Additionally, we learn a shared latent space between human motion and IMU data, incorporating with Zipf’s law regularization to align the frequency distribution of tokens with the inductive bias of natural language.

In order to integrate learned inertial tokens with language vocabulary, it is required to map tokens of multiple modalities into a shared semantic space. However, modern LLMs typically rely on high-dimensional token embeddings (e.g., $3,584$-dimensional text embeddings for Qwen2-7B-Instruct~\cite{yang2024qwen2} model). Therefore, directly learning the discrete latent space of sparse IMU data on such a high dimension is inefficient. To address this challenge, we pretrain a projection layer composed of $8$ Transformer blocks to project low-dimensional inertial tokens onto the LLM's text embedding space. The projected inertial tokens are then concatenated with text tokens and fed into causal language model~\cite{yang2024qwen2} with masks, enabling whole-sequence understanding of inertial tokens. Finally, we fine-tune the language model using LoRA adapters~\cite{hu2022lora} to enhance its flexibility and customization such as acting as a professional fitness coach or nutritionist, delivering tailored feedback on user actions.

To summarize, our main contributions include:
\begin{itemize}
    \item We propose the first multimodal system with real-time motion capture and online motion analysis through sparse IMU signals.
    \item We introduce a novel distribution matching learning strategy and jitter-reduced tokenizer for representing continuous and jittery IMU signals as a sequence of tokens, achieving robust motion capture results under various noisy input conditions.
    \item We integrate jitter-reduced inertial tokens into LLM and enable interaction-friendly applications for real-time motion understanding, including description and instruction with optionally customized styles.
\end{itemize}
\section{Related Work}
\label{sec:rw}

\paragraph{Inertial Posers.}
Motion capture solutions using inertial measurement units (IMUs) have gained significant traction in recent years. Commercial products like Noitom \cite{noitom} and Movella \cite{Movella} leverage dense IMUs to offer high-quality, portable, and real-time applications. However, the usage of these IMU systems can be cumbersome because they require numerous sensors to be attached to the body, which can be inconvenient and intrusive for users. Since the exploration of SIP \cite{von2017SIP}, learning-based methods under the sparse sensor configuration \cite{huang2018DIP,TransPoseSIGGRAPH2021,TIP22,PIPCVPR2022,van2024diffusionposer,yi2024pnp} (called ``inertial poser'') and head-mounted device \cite{du2023avatars,yang2024divatrack,dai2024hmd,starke2024categorical} (called ``three-point tracker'') markedly improved the cost and convenience. These advancements, facilitated by real and synthetic datasets \cite{trumble2017total,huang2018DIP,AMASS:ICCV:2019}, have led to consumer-level products like Mocopi \cite{mocopi}. Despite these developments, inertial methods inherently suffer from issues such as sensor drift and lack of globally positional reference. To mitigate such defects, hybrid approaches \cite{liang2023hybridcap,ren2023lip,EgoLocate2023,pan2023fusing} fuse sparse IMU sensors with monocular vision signals. Furthermore, the recognition and analysis of human motion based on IMUs remain limited to simple classification tasks within some fixed action categories \cite{stromback2020mm,chen2015utd,yan2024language}. Such constraints highlight the need for a more robust ``inertial poser'' and an open-vocabulary system for human motion analysis from IMU signals.

\vspace{-4mm}
\paragraph{Human Motion Understanding with Natural Language.}
Significant strides have been made in the field of human motion understanding with natural language, driven by advanced transformer \cite{vaswani2017attention} and diffusion models \cite{ho2020denoising,song2020denoising}. Specifically, techniques such as text-to-motion generation \cite{tevet2023human,tevet2022motionclip,zhang2024motiondiffuse,Guo_2022_CVPR,zhang2024motiongpt}, motion controlling \cite{delmas2023posefix,athanasiou2024motionfix,xie2024omnicontrol,huang2024controllable}, and motion recognition \cite{locate:2024} learn a conditional probability distribution given textual descriptions and motion sequences respectively. More recent work \cite{jiang2023motiongpt,li2024unimotion,wu2024motionllm,li2024lamp,zhou2024avatargpt,wang2024motiongpt} utilized powerful LLMs to build unified models, enabling versatile motion-language tasks within a single framework. In addition, the skeleton-based action recognition problem also achieved promising progress under the paradigm of fine-tuning LLMs \cite{qu2024llms,mo2024mochat,do2024tdsm}. Despite the impressive performance of these approaches, there still remain two notable issues. Firstly, the motion-language alignment is sub-optimal for understanding and analysis, because the SMPL parametric representation \cite{SMPL2015,MANO:SIGGRAPHASIA:2017,SMPLX2019} is essentially an approximation of real human movements. It simplifies the complex and nuanced nature of human motion, which can fail to express the subtleties and variances of actual inertia. Secondly, to interact with agents using human behavior in the real world, these methods inevitably rely on pose estimators which can introduce uncontrollable noises and errors. These limitations underscore the importance of the alignment between observable and raw sensor signals with natural language.

\begin{figure*}[t]
  \centering
  \includegraphics[width=\linewidth]{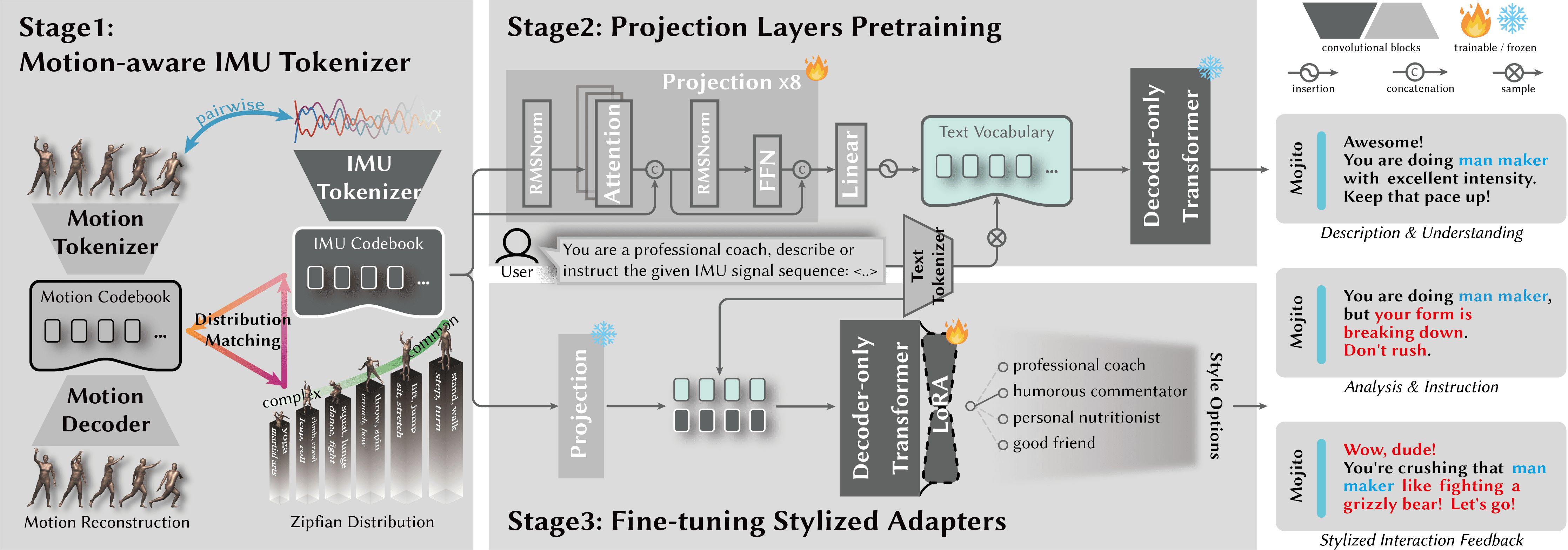}
  \vspace{-20px}
  \captionof{figure}{\textbf{Overview of our training pipeline.} We quantize continuous and jittery IMU signals to a sequence of jitter-reduced and motion-aware inertial tokens by learning a IMU tokenizer through distribution matching strategy and adopt semantic aligned and LoRA fine-tuned LLM to generate precise, professional and stylistic text feedback for human motion analysis.}
\label{fig:train}
\vspace{-10px}
\end{figure*}

\vspace{-4mm}
\paragraph{Human Motion Analysis by Multimodal LLMs.}
In recent years, vision-language models (VLMs) have gained significant traction, demonstrating impressive results in video captioning, reasoning, and understanding tasks \cite{feng2024chatpose,chen2024motionllm,wang2024internvideo2,li2024human,internlmxcomposer,internlmxcomposer2,internlmxcomposer2_4khd,internlmxcomposer2_5}. These models have shown their potential to bridge visual and textual modalities for complex reasoning and semantic understanding \cite{jiang2024motionchain,bao2024exploiting,lin2024chathuman,li2024techcoach,han2023imagebind}. However, when applied to human activity analysis \cite{chen2024internvl2,li2024mvbench,zhang2024narrative}, video data often proves to be a heavy and redundant modality. It introduces substantial amounts of static scenes and irrelevant background information while being susceptible to occlusion, making vision-based approaches inefficient and resource-intensive for human activity analysis. Moreover, state-of-the-art VLMs, including proprietary large language models like GPT-4 and Gemini, fall short of providing real-time analysis and dynamic feedback, which significantly limits their applicability in real-world interactive scenarios. In contrast, we leverage inertial measurement unit (IMU) signals for human motion analysis\cite{li2024sensorllm,moon2022imu2clip,moon2024anymal,girdhar2023imagebind}. IMU signal is a sparse and lightweight modality, captured through body-worn devices, and provides an accurate representation of human actions in the 3D world. Its efficiency and low computational cost make it well-suited for enabling real-time interactive applications, addressing the limitations of vision-based approaches, and expanding the scope of multimodal human activity analysis.
\section{Jitter-reduced IMU Tokenizer}
\label{sec:method1}
In practical application of long-sequence motion capture and online motion analysis using IMU sensors, device connection, signal transmission and wearing fashion can significantly influence the quality of motion capture and the convenience of user experience. However, it is challenging due to the inherent defects of IMUs such as data drifting and jittery signals. Therefore, we start by proposing a jitter-reduced and motion-aware IMU tokenizer to represent sparse inertial signals by discrete tokens, which can compress continuous IMU signals into a fixed collection of latent codes shared with motion latent space. Consequently, the discrete inertial tokens can be integrated into the vocabulary of LLMs, while also support high-quality motion reconstruction. The IMU tokenizer is built upon the standard VQ-VAE framework \cite{van2017neural}, with a novel distribution matching strategy to learn an approximate latent space of corresponding human motion. Additionally, linguistic properties are assigned to the learned inertial tokens through regularization under Zipf's law \cite{zipf2013psycho}, which facilitates following semantic alignment with natural language.

\subsection{Motion VQ-VAE}
\label{sec:motion_vq}
We first follow MotionGPT \cite{jiang2023motiongpt} to learn a VQ-VAE for human motion. Differently, we represent human motion with a complete state of root joint and foot-ground contacts to suit IMU sensor characteristics. In addition, we involve regularization terms on foot-ground contacts to eliminate jittery results and sliding artifacts in decoded motion.

\vspace{-4mm}
\paragraph{Motion Representation.} While HumanML3D \cite{Guo_2022_CVPR} establishes an effective motion representation for text-to-motion generation tasks, it is limited to incomplete global dynamics and missing foot-ground contacts. Inspired by HuMoR \cite{rempe2021humor}, we incorporate root translation and angular velocity along all three axes into our representation to improve expressiveness. Specifically, we represent a motion sequence as
\begin{equation}
    \mathbf M^{1:T} = \left[
        \mathbf r
        \quad \dot{\mathbf{r}}
        \quad \mathbf \Phi 
        \quad \dot{\mathbf{\Phi}}
        \quad \mathbf{j}^r
        \quad \mathbf{j}^p
        \quad \mathbf{j}^v
        \quad \mathbf{p}
    \right] \in \mathbb{R}^{T\times d_m}\text{,}
\end{equation}
where $T$ is the sequence length. Within the representation, we first include the root translation $\mathbf r\in\mathbb R^{T\times3}$, linear velocity $\dot{\mathbf{r}} \in\mathbb R^{T\times3}$, orientation $\mathbf \Phi\in\mathbb R^{T\times6}$, and angular velocity $\dot \Phi\in\mathbb R^{T\times 3}$. Then, we use $\mathbf {j}^r\in \mathbb R^{T\times 6J}$, $\mathbf {j}^p\in\mathbb R^{T\times 3J}$, $\mathbf {j}^v\in\mathbb R^{T\times3J}$ to represent local joint rotations, positions, and velocities, respectively. Finally, $\mathbf p\in \mathbb R^{T\times 4}$ records the binary contact labels of toes and heels. Here, $d_m=271$ is the dimension of our motion representation, and $J=21$ is the number of local joints. All the rotational parts are in 6D rotation convention \cite{zhou2019continuity}.

\vspace{-4mm}
\paragraph{Training of Motion VQ-VAE} Given a motion sequence $\mathbf{M}^{1:T}$, we first encode it into discrete latent codes $\mathbf{Z}^{\text{motion}} \in \mathbb{R}^{S\times d_z}$ using 1D convolution layers, where $d_z=512$ is the dimension of latent code, and $S$ is the number of the resulting latent codes. We define the hyperparameter $l=\lfloor T/S\rfloor$ as the compression rate for discretization. Following the encoding process, each latent code $\mathbf{z}_s^{\text{motion}}$ is quantized to a learned codebook $\mathbf{C}^{\text{motion}}\in \mathbb{R}^{K\times d_z}$, where $K$ is the codebook size. The quantization process runs as follows
\begin{equation}
    \mathbf{b}_{s}^{\text{motion}} = \underset{\mathbf{c}_k^{\text{motion}}}{\argmin}\left\Vert \mathbf{z}_{s}^{\text{motion}}-\mathbf{c}_k^{\text{motion}} \right\Vert_2^2\text{,}
\end{equation}
which selects the nearest codebook entry according to Euclidean metric, and results in the motion token sequence $\mathbf B^{\text{motion}}\in \mathbb{R}^{S\times d_z}$. Subsequently, $\mathbf{B}^{\text{motion}}$ is fed into the decoder to reconstruct original motion sequence $\hat{\mathbf M}^{1:T'}$ with possible truncation $T'=lS$. To train the motion VQ-VAE, we utilize the discrete representation learning objective \cite{van2017neural} to supervise our network
\begin{equation}
    \mathcal{L}_{\text{vq}} = \lambda_{\text{recon}}\mathcal{L}_{\text{recon}} + \lambda_{\text{commit}}\mathcal{L}_{\text{commit}}\text{.}
\end{equation}
Specifically, the reconstruction loss is defined as
\begin{equation}
    \mathcal L_{\textrm{recon}} = \frac{1}{T'}\left\Vert \hat{\mathbf M}^{1:T'} - \mathbf M^{1:T'} \right\Vert_2^2\text{,}
    \label{eq:recon_loss}
\end{equation}
and the commit loss with gradient pass-through is
\begin{equation}
    \mathcal L_{\textrm{commit}} = \frac{1}{S} \left\Vert\mathbf{Z}^{\textrm{motion}} - \mathbf{B}^{\textrm{motion}}\right\Vert_2^2\text{.}
    \label{eq:commit_loss}
\end{equation}
\begin{figure}[t]
  \centering
  \includegraphics[width=\linewidth]{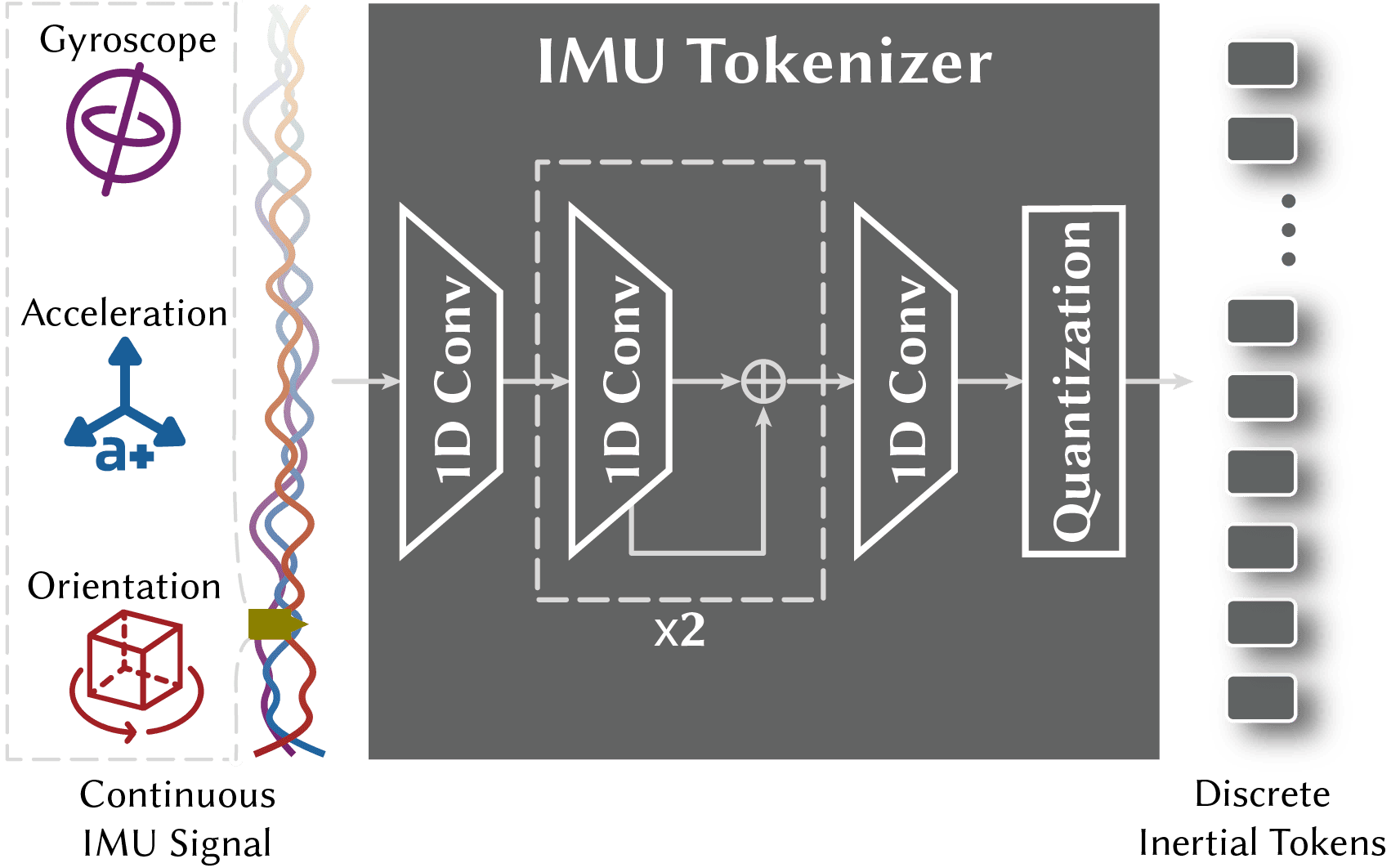}
  \vspace{-15px}
  \captionof{figure}{\textbf{IMU Tokenizing Process.} The rotation, acceleration, and angular velocity components of the IMU signal are first flattened and concatenated. The resulting sequence is then processed by an encoder comprising multiple 1D convolutional layers and subsequently passed through a quantizer to generate the jitter reduced inertial tokens.}
\label{fig:imu_tokenizer}
\vspace{-10px}
\end{figure}
Additionally, to improve the fidelity of reconstructed motion and eliminate the foot-ground sliding artifacts, we follow HuMoR \cite{rempe2021humor} to constrain the foot-ground interactions using
\begin{equation}
    \mathcal{L}_{\text{foot}} = \lambda_{\textrm{contact}}\mathcal L_{\textrm{contact}} + \lambda_{\textrm{slide}}\mathcal L_{\textrm{slide}}\text{,}
\end{equation}
where the contact discrimination loss is
\begin{footnotesize}
\begin{equation}
    \mathcal L_{\textrm{contact}} = \frac{1}{T'}\sum_{i\in\left\{1,2,3,4\right\}} \left[-\mathbf{p}_i\log\hat{\mathbf{p}}_i-\left(1-\mathbf{p}_i\right)\log\left(1-\hat{\mathbf{p}}_i\right)\right] \text{,}
    \label{eq:contat_loss}
\end{equation}
\end{footnotesize}
and the sliding penalty is
\begin{equation}
    \mathcal L_{\textrm{slide}} = \frac{1}{T'}\sum_{i\in\{1,2,3,4\}}\hat{\mathbf{p}}_i\left\Vert \mathbf j^v_{\operatorname{foot}\left(i\right)}\right\Vert_2^2\text{.}
    \label{eq:velocity_loss}
\end{equation}
Overall, the total training loss of our motion VQ-VAE is
\begin{equation}
    \mathcal{L}_{\text{motion}} = \mathcal{L}_\text{vq} + \mathcal{L}_\text{foot}\text{.}
\end{equation}
For following distribution matching, we maintain the frequency distribution of the motion codebook within each training batch
\begin{footnotesize}
\begin{equation}
  \mathbf{F}^{\textrm{motion}} = \frac{1}{S}\pi\left(\sum_{s=1}^S \mathcal{G}\left(\left\{-\left\Vert\mathbf{z}_s^\textrm{motion}-\mathbf{c}_k^\textrm{motion}\right\Vert_2^2\right\}_{k=1}^K\right)\right)\text{,}
\label{eq:gumbel_softmax}
\end{equation}
\end{footnotesize}
where $\mathcal{G}(\cdot):\mathbb{R}^{K}\mapsto\mathbb{R}^{K}$ is the differentiable sampling procedure using Gumbel-Softmax trick \cite{jang2016categorical}, and $\pi(\cdot)$ operates sorting on the token frequencies in descending order.

\subsection{IMU Tokenizer}
\label{sec:imu_vq}
In this subsection, we introduce the jitter-reduced and motion-aware IMU tokenizer. To facilitate the integration of continuous inertial signals with natural language in a manner compatible with large language models (LLMs), we propose a novel approach that encodes inertial signals into discrete tokens, as shown in Fig.~\ref{fig:imu_tokenizer}. These tokens are designed to align seamlessly with the LLM vocabulary, enabling direct incorporation into the language modeling framework. Meanwhile, to empower inertial tokens with the capability of reproducing high-quality 3D motion, we devise a novel distribution matching strategy to approximate the corresponding motion latent space. Therefore, the learned IMU codebook can be utilized for motion reconstruction and analysis.

\vspace{-4mm}
\paragraph{Inertia Representation} In prior works \cite{huang2018DIP,TransPoseSIGGRAPH2021,TIP22,PIPCVPR2022,yi2024pnp}, inertial signals are considered as the composition of orientation and linear acceleration. However, to fully utilize the sensor measurements of the accelerometer, gyroscope, and magnetometer, we represent an inertia sequence as follows
\begin{equation}
    \mathbf I^{1:T} = [\mathbf{q} \quad \mathbf{a} \quad \bm{\omega}] \in\mathbb R^{T\times d_u}\text{,}
\label{eq:inertia_representation}
\end{equation}
which includes orientation $\mathbf q\in\mathbb R^{T\times 6N}$, free acceleration $\mathbf a\in\mathbb R^{T\times 3N}$ and angular velocity $\bm{\omega} \in \mathbb{R}^{T\times 3N}$. In this work, we utilize a configuration of $N=6$ IMU sensors. The collected inertial data is represented in the feature space with a dimensionality of $d_u=72$, capturing comprehensive motion characteristics.

\vspace{-4mm}
\paragraph{Data Pre-processing} Due to the scarcity of MoCap data paired with real IMU readings \cite{huang2018DIP,trumble2017total,dai2024hmd}, we simulate synthetic IMU signals on extensive motion data \cite{huang2018DIP,TransPoseSIGGRAPH2021}. To model the characteristics of IMU sensors, such as data drift, we follow PNP \cite{yi2024pnp} to use random walk variables to mimic cumulative error. Since acceleration data can fluctuate violently within a wide range, we normalize it to a standard normal distribution using the mean and variance determined on the training dataset. This preprocessing procedure mitigates the impact of high-frequency noise spikes and irregular waves while preserving the drifting feature, improving the learning stability of the tokenizer.

\vspace{-4mm}
\paragraph{Training of IMU Tokenizer} Given an inertia sequence $\mathbf{I}^{1:T}$, we learn to construct a codebook $\mathbf{C}^{\text{imu}} \in \mathbb{R}^{K\times d_z}$. To be specific, each codebook entry $\mathbf{c}_k^{\text{imu}}$ is updated through exponential moving average (EMA) according to \cite{razavi2019generating}
\begin{align}
    \mathbf{c}_k^{\textrm{imu}} &\leftarrow \frac{\bm{\sigma}_k}{\delta_k} \notag\\
    \bm{\sigma}_k &\leftarrow \gamma\bm{\sigma}_k + \left(1-\gamma\right)\sum_{s=1}^S
    \mathbbm{1}\left(\mathbf{b}_s^{\textrm{imu}}=\mathbf{c}_k^{\textrm{imu}}\right)\mathbf{z}_s^{\textrm{imu}} \notag\\
    \delta_k &\leftarrow \gamma\delta_k + \left(1-\gamma\right)\sum_{s=1}^S\mathbbm{1}\left(\mathbf{b}_s^{\textrm{imu}}=\mathbf{c}_k^{\textrm{imu}}\right)\text{,}
\label{eq:codebook_update}
\end{align}
where the summation $\sum_{s=1}^S\mathbbm{1}\left(\mathbf{b}_s^{\textrm{imu}}=\mathbf{c}_k^{\textrm{imu}}\right)$ records the count that $\mathbf{c}_k^{\textrm{imu}}$ is selected. Similar to Eq.\ref{eq:gumbel_softmax}, we also maintain the frequency distribution of IMU codebook $\mathbf{F}^{\text{imu}}$ within each training batch. To inject motion dynamics and inductive bias of natural language into inertial tokens, we propose to learn by unsupervised distribution matching, inspired by CM \cite{starke2024categorical}. Specifically, the training objective of our motion-aware IMU tokenizer is
\begin{equation}
    \mathcal{L}_{\text{imu}} = \lambda_{\text{code}}\mathcal{L}_{\text{code}} + \lambda_{\text{dist}}\mathcal{L}_{\text{dist}}\text{,}
\label{eq:loss_imu_tokenizer}
\end{equation}
where the code matching loss enforces the quantized token from the IMU tokenizer close to that from the motion tokenizer
\begin{equation}
    \mathcal L_{\text{code}} = \frac{1}{S}\left\Vert \mathbf{B}^{\text{imu}}-\mathbf{B}^{\text{motion}}\right\Vert_2^2\text{.}
\label{eq:token_match_mse}
\end{equation}
We also incorporate Zipf's law \cite{zipf2013psycho,piantadosi2014zipf}, a principle about the word frequency distribution in natural language, as a regularization term to enhance the linguistic properties of the inertial tokens \cite{qu2024llms,papadimitriou2023pretrain}. Formally, the Zipfian distribution $\mathbf{F}^{\text{zipf}}$ is defined as
\begin{equation}
    \mathbf{F}^{\text{zipf}} \propto \left\{\frac{1}{(k+\beta)^\alpha} \;\middle|\; k\in\left\{1\dots K\right\}\right\}\text{,}
\label{eq:zipf_law}
\end{equation}
where $\alpha\approx 1$, $\beta\approx2.7$, and the distribution matching loss tries to minimize the Jensen-Shannon (JS) divergence between the categorical frequency distribution of IMU and motion codebook
\begin{small}
\begin{equation}
    \mathcal L_{\text{dist}} = \operatorname{JS}\left({\mathbf{F}^\text{imu}} \;\middle|\middle|\; {\mathbf{F}^{\text{motion}}}\right) + \lambda_{\text{zipf}}\operatorname{JS}\left({\mathbf{F}^\text{motion}} \;\middle|\middle|\; {\mathbf{F}^{\text{zipf}}}\right)\text{.}
\label{eq:distribution_matching}
\end{equation}
\end{small}

\subsection{Implementation Details}
\label{sec:implementation_detail_part1}
To accommodate the high frame rates typical of IMU sensors, we standardize motion data across various datasets to 50 and 60 frames per second (fps). Consequently, we adopt a codebook size of $K=1024$, which exceeds that used in MotionGPT \cite{jiang2023motiongpt}, and trained on 20 fps data to better capture the increased temporal resolution. We observed that higher compression rates $l$ can introduce square-wave-like artifacts in the encoded IMU signals in our experiment. To address this, we set $l=4$, achieving a balanced trade-off between the compactness and the expressiveness of the discrete token representation. During training, the EMA coefficient is $\gamma=0.99$, and the loss weights are configured as: $\lambda_{\textrm{recon}}=1.0$, $\lambda_{\textrm{commit}}=0.02$, $\lambda_{\textrm{contact}}=0.01$, $\lambda_{\textrm{slide}}=0.01$, $\lambda_{\textrm{dist}}=1.0$, $\lambda_{\textrm{code}}=1.0$, and $\lambda_{\textrm{zipf}}=0.2$.
We utilize the AdamW optimizer \cite{loshchilov2017decoupled} with learning rate $\textrm{lr}=2\times10^{-4}$ and cosine annealing scheduler \cite{loshchilov2016sgdr}. The training batch size is set to $512$ for both motion and IMU tokenizer.

\section{Language Model with Inertial Tokens}
\label{sec:method2}


Using the aforementioned IMU tokenizer, our method discretizes continuous and jittery IMU signals into sequential jitter-reduced tokens. However, unlike the high-dimensional embedding space of LLMs, the learned inertial tokens reside in a compact, low-dimensional latent space. Consequently, it is essential to pre-align these inertial tokens with language embeddings to facilitate subsequent multimodal understanding. In this section, we first introduce our method for generating curated textual annotations paired with inertial and motion sequences. (Sec.~\ref{sec:data_preparation}). Following this, we detail our method for projecting the inertial tokens into the vocabulary space of Qwen2-7B-Instruct language model \cite{yang2024qwen2} in Sec.~\ref{sec:pretrain}, and introduce our LoRA model adapters, which enhance the system's professionalism, rationality, and stylization in Sec.~\ref{sec:finetuning_lora}

\begin{figure}[t!]
  \centering
  \includegraphics[width=\linewidth]{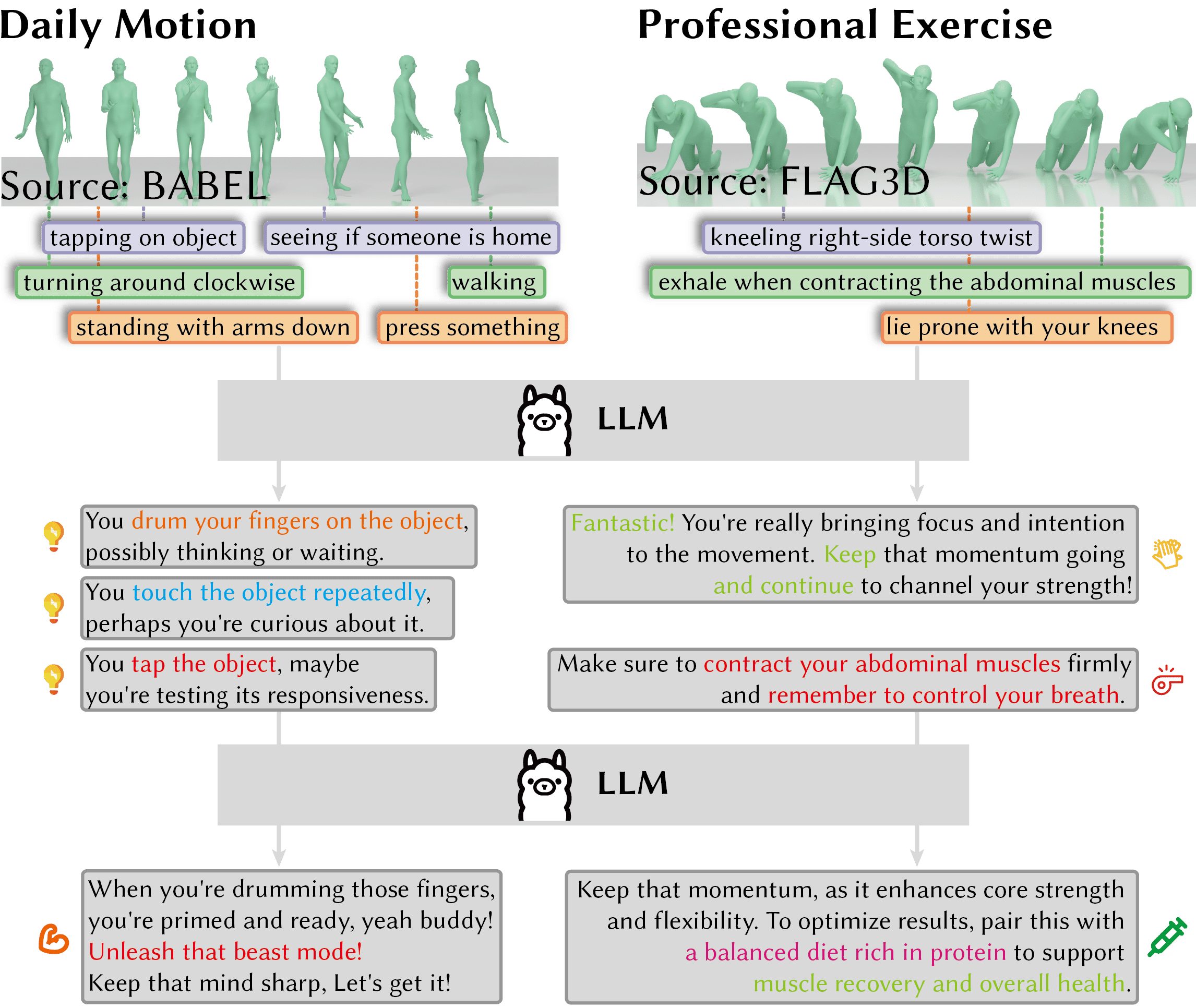}
  \vspace{-20px}
  \captionof{figure}{\textbf{Data Generation Pipeline.} The corresponding motion label is first extracted and expanded into a descriptive sentence using the LLM. Subsequently, a prompt is employed to generate a more refined and professional description or instructional output.}
\label{fig:data}
\vspace{-10px}
\end{figure}

\subsection{Data Preparation}
\label{sec:data_preparation}

To prepare extensive training data, we instruct GPT-4o-mini with carefully designed prompts to automatically rephrase raw textual annotations from original datasets or generate interactive dialogues based on concise action labels. Given the rich diversity of human motion, as illustrated in Fig.~\ref{fig:data}, we categorize the collected datasets into two broad groups: ``daily motion'' and ``professional exercise'', and annotate them with descriptions and instructions respectively.

\vspace{-4mm}
\paragraph{Descriptive and Instructive Text Data Generation} 

For the daily motion category, we prompt GPT-4o-mini to generate responses structured into two parts: an objective description of the given action and a subjective analysis of the motivation or intention behind the human actor's behavior. For the instructive category, we request detailed motion analysis, supplemented with assessments and instructions, including encouragements or critiques expressed with clear attitudes. To ensure the feedback resembles human-like communication, we constrain the generated texts to adopt a second-person narrative style.

\vspace{-4mm}
\paragraph{Stylized Feedback Generation} 

To further diversify and enrich the dataset, we introduce stylized roles that incorporate distinct personalities and linguistic tones. First, we define each role by specifying its personality traits, typical phrasing and overarching stylistic characteristics. Subsequently, we either select an existing role-specific utterance or virtually construct one as an exemplar. This approach generates vivid, character-driven dialogues that encompass a wide range of stylistic variations, enhancing the diversity of interactions within our system.

\begin{figure}[t!]
  \centering
  \includegraphics[width=\linewidth]{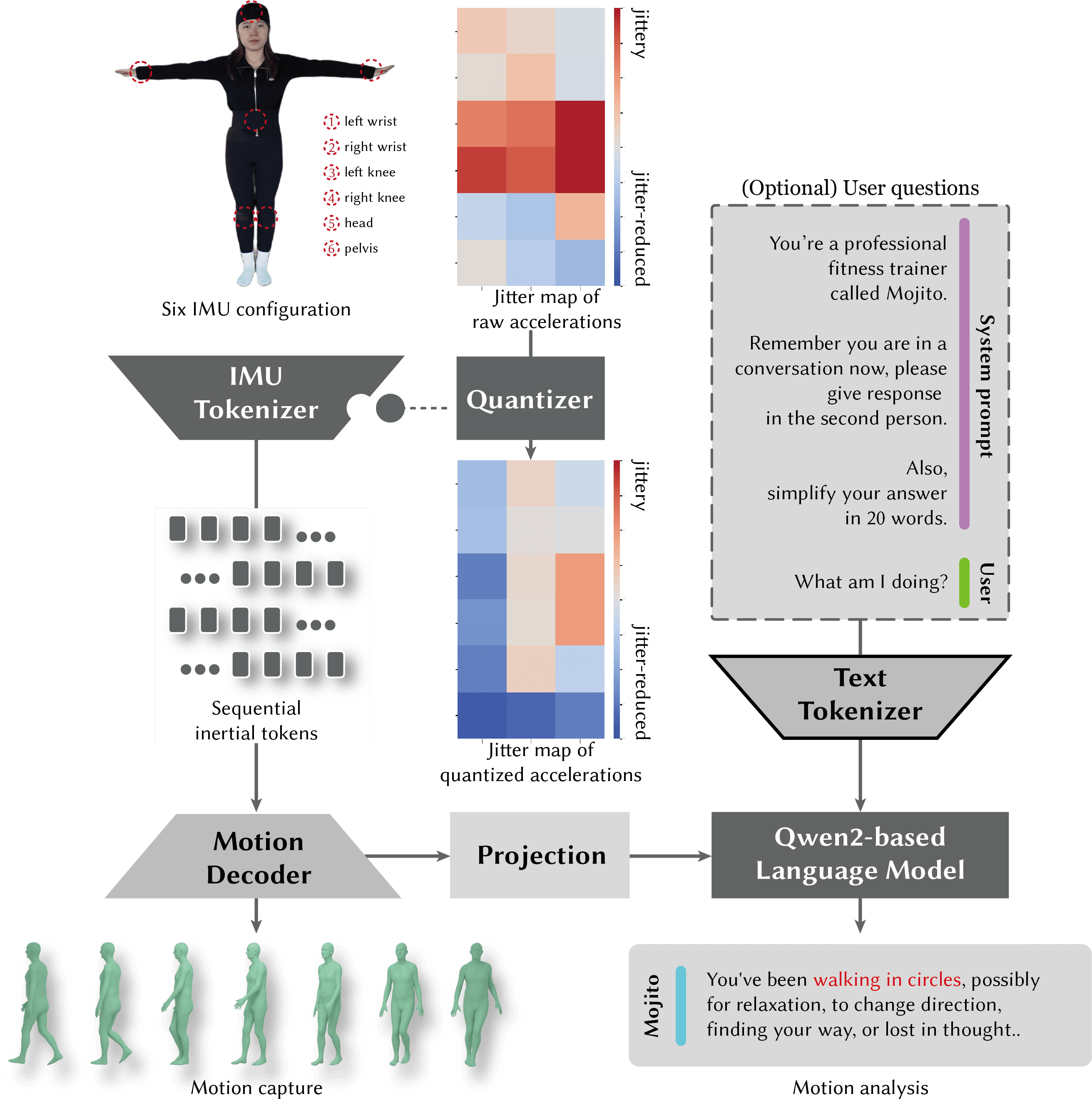}
  \vspace{-20px}
  \captionof{figure}{\textbf{Inference Pipeline.}
    Jittery IMU signals are first tokenized into jitter-reduced inertial tokens. These tokens are concurrently processed in two ways: (1) they are decoded by the learned motion decoder to reconstruct human motion, and (2) they are projected into the language semantic space via the pretrained projection module for motion analysis.  
  }
\label{fig:inference_pipeline}
\vspace{-10px}
\end{figure}

\subsection{Projecting Inertial Token to Text Embedding}
\label{sec:pretrain}

Seamless translation and understanding between IMU and textual modalities necessitate a shared and well-aligned embedding space. However, achieving this is challenging due to the significant disparity in embedding dimensions between well-known LLMs and inertial tokens. For instance, the Qwen2-7B-Instruct~\cite{yang2024qwen2} employs a text embedding dimension $d_h=3584$, which is substantially larger than $d_z=512$ dimension of our inertial tokens. To address this issue, we adopt a strategy inspired by OneLLM~\cite{han2024onellm}, introducing a projection module $\mathcal{P}_{\theta}$ that maps inertial tokens $\mathbf{b}_s^{\textrm{imu}}$ into the text embedding space.

\vspace{-4mm}
\paragraph{Training of Projection Module} 

As illustrated in Fig.~\ref{fig:train}, our projection module comprises eight transformer blocks followed by a linear layer. Each transformer block incorporates a self-attention layer, a feed forward network, and skip connections, following the architecture of Llama3~\cite{dubey2024llama}, to ensure effective gradient flow. For each inertial token $\mathbf{b}_s^{\textrm{imu}}$, the projection module maps it to the text embedding space of Qwen2-7B-Instruct~\cite{yang2024qwen2}:
\[
\mathbf{e}_s = \mathcal{P}_{\theta}(\mathbf{b}_s^{\textrm{imu}}) \in \mathbb{R}^{d_h}\text{.}
\]
Concurrently, optional user-provided textual prompts are tokenized and embedded into the same space, after which they are concatenated with projected inertial tokens to form the input for the language model. At this stage, we keep the Qwen2-7B-Instruct~\cite{yang2024qwen2} model frozen and train only the projection layer $\mathcal{P}_{\theta}$ using cross-entropy loss. To establish semantic associations between inertial tokens and text, rather than focusing on the intrinsic causality within the inertial token sequence, we employ a training strategy inspired by chatting language models. This involves augmenting mask to all input tokens and excluding them from the loss computation during training.

\subsection{Fine-tuning Language Model Adapters}
\label{sec:finetuning_lora}


To further empower our language model with greater flexibility and customization capabilities, we finetune 4 Low-Rank Adaptation (LoRA)~\cite{hu2022lora} adapters. These adapters enable the generation of stylized feedback with character-specific tones, allowing for tailored responses in customized roles. During fine-tuning, both the projection module $\mathcal{P}_{\theta}$ and the pretrained weights of the Qwen2-7B-Instruct~\cite{yang2024qwen2} language model remain frozen, with updates applied exclusively to the LoRA adapters. This fine-tuning process enriches the model's linguistic understanding of the same inertial sequences, facilitating personalized and adaptable usage scenarios.
\section{Experiments}
\label{sec:experiment}

\begin{figure*}[t!]
    \centering
    \includegraphics[width=0.87\linewidth]{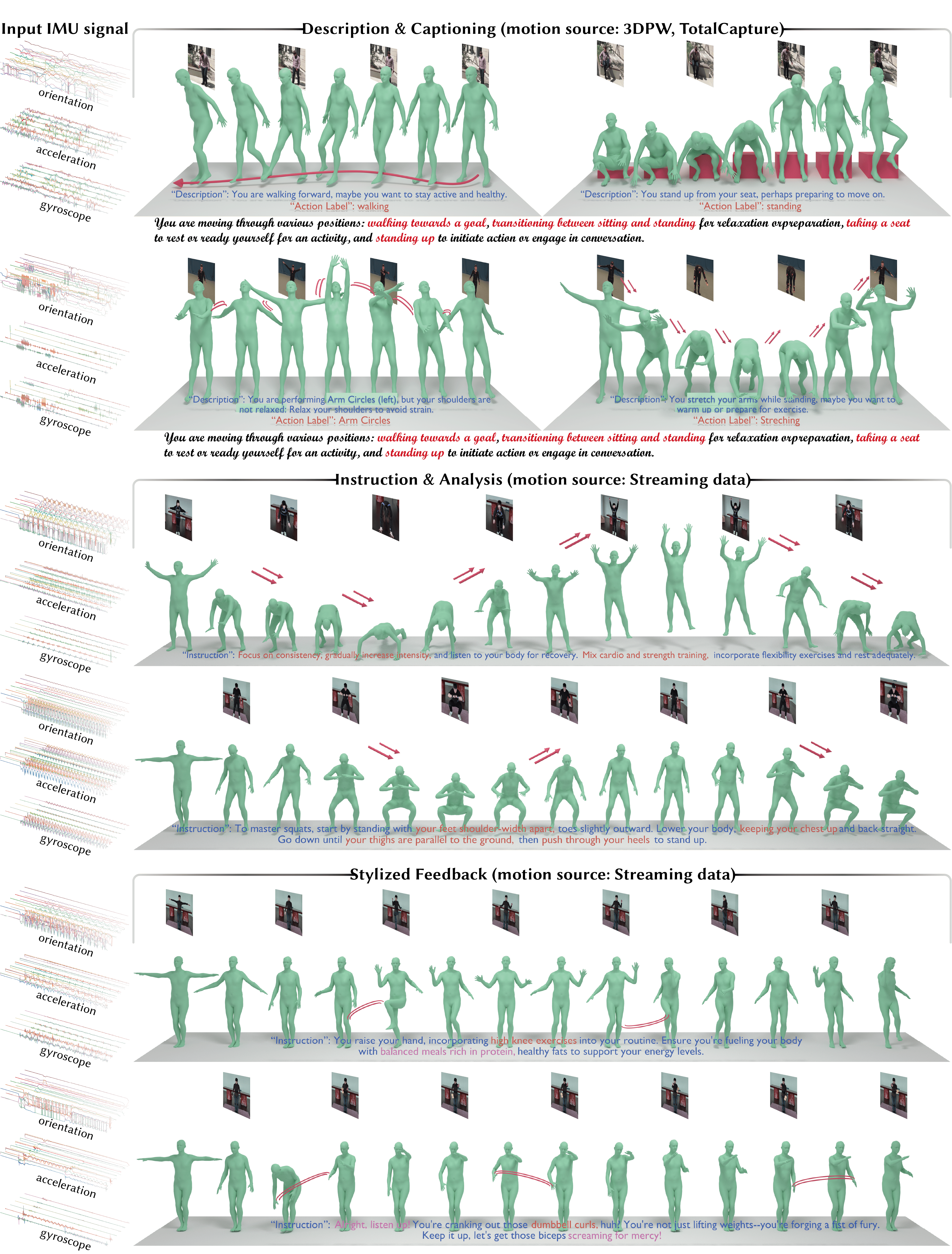}
    \caption{\textbf{Results Gallery.} We present input IMU signals, MoCap results, system analysis, and RGB references.}
\label{fig:gallery}
\end{figure*}

\begin{figure*}[t!]
    \centering
    \includegraphics[width=\linewidth]{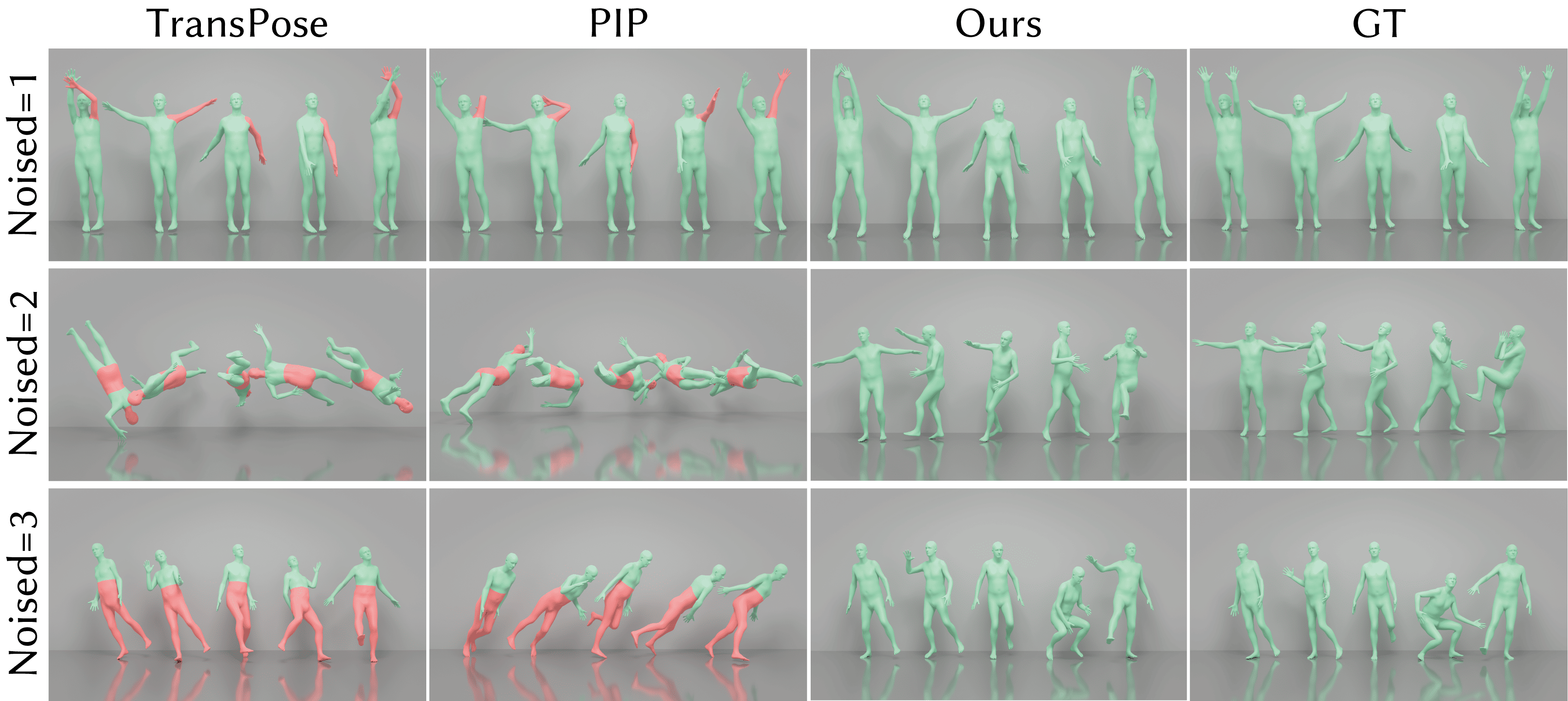}
    \caption{\textbf{Qualitative Comparisons of Motion Reconstruction.}
    We evaluate our method on the TotalCapture~\cite{trumble2017total} and PICO-FreeDancing~\cite{dai2024hmd} dataset. IMU sensors attached to the highlighted body parts (shown in red) are disturbed by noises. The first row demonstrates the single-sensor noised condition, where noise is applied to the left wrist sensor. The second row presents the two-sensor noised condition, with noise introduced to the pelvis and head sensors. The third row depicts the three-sensor noised condition, where noise affects the pelvis and both knee sensors, while the head and wrist sensors remain unaffected. This configuration corresponds to the three-point tracking setup commonly used in VR systems.
    }
\label{fig:mocap_comparison}
\end{figure*}

\begin{table*}[t!]
    \centering
    \tabcolsep=0.15cm
    \tiny
    \begin{tabular}{lcccccccccc}
        \toprule
        & \multicolumn{3}{c}{Noised=1 (MPJPE/Mesh Err/Jitter) $\downarrow$} & \multicolumn{3}{c}{Noised=2 (MPJPE/Mesh Err/Jitter) $\downarrow$} & \multicolumn{3}{c}{Noised=3 (MPJPE/Mesh Err/Jitter) $\downarrow$}\\
        \cmidrule(lr){2-4}\cmidrule(lr){5-7}\cmidrule(lr){8-10}
        Methods & BABEL & TotalCapture & Pico-FreeDancing & BABEL & TotalCapture & Pico-FreeDancing & BABEL & TotalCapture & Pico-FreeDancing\\
        \midrule
        TransPose 
        & 15.20/18.07/1387.71 & 16.95/20.08/1285.59 & 18.89/23.93/1401.72 
        & 22.76/27.63/2070.31 & 23.67/28.68/1913.04 & 24.87/30.67/2092.98
        & 36.27/42.19/3960.68 & 37.02/43.31/4930.36 & 37.66/44.34/3979.20\\
        PIP 
        & 16.13/19.89/129.77 & 26.25/32.32/200.70 & 11.96/17.53/27.57$^*$
        & 15.34/19.69/133.95 & 32.54/45.31/374.82 & 13.87/22.12/51.38$^*$
        & 33.81/40.43/367.92 & 32.62/45.23/510.24$^*$ & 33.43/46.55/480.17$^*$\\
        Ours & 
        \textbf{10.44}/\textbf{13.63}/\textbf{1.40} & 
        \textbf{12.27}/\textbf{15.37}/\textbf{1.46} & 
        \textbf{11.69}/\textbf{16.59}/\textbf{1.04} &
        \textbf{12.81}/\textbf{16.97}/\textbf{1.47} &
        \textbf{14.30}/\textbf{18.54}/\textbf{1.52} &
        \textbf{13.54}/\textbf{19.14}/\textbf{1.11} &
        \textbf{16.05}/\textbf{21.41}/\textbf{1.86} &
        \textbf{17.19}/\textbf{22.54}/\textbf{2.18} &
        \textbf{16.73}/\textbf{22.45}/\textbf{1.38} &
        \\
        \bottomrule
    \end{tabular}
    \caption{\textbf{Accuracy On Different Noised Levels.} For the single-sensor noised condition, noise was independently applied to each of the six IMU signals, and the results were averaged. For the two- and three-sensor noised conditions, random sensor combinations were selected, and the results were averaged accordingly. 
    Note that the symbol ``*'' indicates that PIP failed to solve all noisy sequences due to the numerical instability of the physical solver. As a result, these failed sequences were excluded from evaluation.
    }
\label{tab:noised_mocap_comparison}
\end{table*}

In this section, we first introduce motion datasets containing various modalities that we utilize for training and evaluation in Sec.~\ref{sec:datasets}. Subsequently, in Sec.~\ref{sec:mocap_comparison}, we conduct extensive qualitative and quantitative comparison experiments with other state-of-the-art inertial posers to demonstrate the robustness of our method under various noisy environments, which is benefited from jitter-reduced inertial tokens. 
We further compare the precision, brevity, naturalism and professionalism of the textual feedback generated by our method against a well-constructed baseline method based on TransPose~\cite{TransPoseSIGGRAPH2021} and MotionGPT~\cite{jiang2023motiongpt}, as well as other prominent vision-language models. Additional results related to motion reconstruction, accompanied by text descriptions and instructions, are provided in Fig.~\ref{fig:gallery}.

\subsection{Datasets and Evaluation Metrics}
\label{sec:datasets}

\paragraph{Datasets.} We utilize a diverse collection of datasets encompassing motion-only data, real IMU recordings, and textual annotations. Specifically, our motion-aware and jitter-reduced IMU tokenizer is trained on 3DPW~\cite{vonMarcard2018}, Human3.6M \cite{h36m_pami}, TotalCapture \cite{trumble2017total}, PICO-FreeDancing \cite{dai2024hmd}, BABEL \cite{AMASS:ICCV:2019,BABEL:CVPR:2021}, Motion-X++ \cite{lin2024motion}, Fit3D \cite{Fieraru_2021_CVPR}, FLAG3D \cite{tang2023flag3d}, MOYO \cite{tripathi2023ipman}, and EC3D \cite{zhao2022exercise}. For training our language model, we re-formulate textual annotations of BABEL \cite{AMASS:ICCV:2019,BABEL:CVPR:2021} and Motion-X++ \cite{lin2024motion} on daily motions, and generate interactive dialogues based on Fit3D \cite{Fieraru_2021_CVPR}, FLAG3D \cite{tang2023flag3d}, MOYO \cite{tripathi2023ipman}, and EC3D \cite{zhao2022exercise} through the preparation approach mentioned in Sec.~\ref{sec:data_preparation}.

\vspace{-4mm}
\paragraph{Evaluation Metrics.} 
In the following experiments, we evaluate our method's capability in robust motion capture using comprehensive metrics, including:
\begin{itemize}[leftmargin=*]
    \item \textbf{MPJPE (cm)} measures the average Euclidean distance between reconstructed and ground-truth joint positions.
    \item \textbf{Mesh Error (cm)} evaluates the mean Euclidean error of the reconstructed SMPL mesh~\cite{SMPL2015} across all vertices.
    \item \textbf{Jitter ($\bm{10^2}\textrm{m/s}\bm{^3}$)} computes the third derivative of joint positions over time to assess motion smoothness and reasonability.
\end{itemize}

To evaluate the precision and professionalism of the textual feedback generated by our method, we calculate the following two widely used metrics:
\begin{itemize}[leftmargin=*]
    \item \textbf{BERTScore} measures semantic similarity by computing the average consine similarity of contextual embeddings derived from models such as BERT.
    \item \textbf{METEOR} assesses similarity through stem matching, synonym matching, and positional penalties, while also evaluating fluency.
\end{itemize}

\subsection{Evaluating Robustness of Motion Capture}
\label{sec:mocap_comparison}
Inertial poser methods based on recurrent neural networks (RNN) and physics solvers have consistently faced significant challenges related to noise sensitivity. IMU signals are highly susceptible to noises introduced by various uncontrolled factors, such as magnetic fields interference, prolonged usage, and sub-optimal sensor placement. As a result, robustness remains a crucial yet unresolved issue for real-world applications.
The primary limitation of previous works~\cite{huang2018DIP,TransPoseSIGGRAPH2021,PIPCVPR2022,TIP22,yi2024pnp} lies in their reliance on data-driven neural networks trained to map IMU signals directly to human motion through continuous functions. Consequently, when IMU signals contain outliers, these methods fail to effectively eliminate noise, leading to the propagation of signal artifacts into the reconstructed motion results.
In contrast, our method addresses this issue by tokenizing continuous IMU signals into discrete tokens through a motion-aware and jitter-reduced IMU tokenizer. The quantization operation within this process effectively mitigates various noisy conditions, enabling the system to filter out irregularities and even tolerate severely corrupted IMU signals.
To comprehensively evaluate the robustness of our motion capture, we simulate multiple levels of noisy input conditions by adding random noises to the orientation, acceleration, and gyroscope data of different combinations of IMU configurations. Based on these simulated noisy inputs, we qualitatively and quantitatively demonstrate the advantages of our method over existing approaches.

\begin{table}[t!]
    \centering
    \tabcolsep=0.18cm
    \footnotesize
    \begin{tabular}{lccccc}
        \toprule
        & \multicolumn{2}{c}{BERT $\uparrow$} & \multicolumn{2}{c}{METEOR $\uparrow$}\\
        \cmidrule(lr){2-3}\cmidrule(lr){4-5}
        Methods & Descriptive & Instructive & Descriptive & Instructive\\
        \midrule
        Baseline & 0.8603 & 0.8483 & 0.0706 & 0.0746\\
        InternVideo2 & 0.8467 & 0.8454 & 0.0551 & 0.0583\\
        MotionLLM & \cellcolor{pink}{\textbf{0.9085}} & \cellcolor{cyan!30}{0.8622} & \cellcolor{pink}{\textbf{0.3912}} & \cellcolor{cyan!30}{0.1218}\\
        Ours & \cellcolor{cyan!30}{0.8781} & \cellcolor{pink}{\textbf{0.8667}} & \cellcolor{cyan!30}{0.1205} & \cellcolor{pink}{\textbf{0.1510}}\\
        \bottomrule
    \end{tabular}
    \caption{\textbf{Quantitative Comparison on textual feedback accuracy.} 
    Quantative comparison across four methods (Baseline, InternVideo2, MotionLLM, and ours) on two key metrics (BERT, METEOR). The results are presented under two categories, "Descriptive" and "Instructive", revealing that our approach outperforms the baseline and InternVideo2 while achieving performance comparable to VLM-based MotionLLM in generating contextually rich textual feedback.}
\label{tab:analysis}
\end{table}

\vspace{-4mm}
\paragraph{Qualitative Results} As illustrated in Fig.~\ref{fig:mocap_comparison}, our method consistently outperforms prior works \cite{TransPoseSIGGRAPH2021,PIPCVPR2022} across various noise levels and configurations. In cases where noised signals affect specific body part (e.g., the left arm highlighted in red), RNN-based and physics solver-based methods frequently generate inaccurate motions. This is attributed to the continuous function mapping learned by their networks, which are inherently sensitive to input outliers. In contrast, the quantization step in our jitter-reduced IMU tokenizer effectively filters out corrupted parts in inputs, enabling accurate motion reconstruction even under imperfect IMU signals. Notably, our method demonstrates robust performance even when the IMU sensor attached to the pelvis is absent. While existing methods, which rely on root-relative input data representations, tend to generate messy motions when global orientation is invalid, our method effectively leverages valid signals from other sensors and quantizes them into reasonable discrete latent features. This capability significantly enhances the robustness of inertial posers in such challenging scenarios.

\vspace{-4mm}
\paragraph{Quantitative Results}
In addition to qualitative comparisons, we also quantitatively evaluate the motion capture accuracy and robustness under various noisy input configurations. As reported in Tab.~\ref{tab:noised_mocap_comparison}, we present performance of our method and two state-of-the-art approaches~\cite{TransPoseSIGGRAPH2021,PIPCVPR2022} on both synthesized IMU data from BABEL~\cite{braams:babel,AMASS:ICCV:2019} and real IMU recordings from TotalCapture~\cite{trumble2017total} and PICO-FreeDancing~\cite{dai2024hmd}. Our method outperforms other approaches by a large margin in both motion capture accuracy and smoothness. In particular, under severe noise conditions, our method maintains stable performance, achieving 2 times lower MPJPE and hundreds times lower jitters.

\subsection{Evaluating Quality of Textual Feedback}
We further evaluate the precision and professionalism of the motion analysis generated by our method. To establish a meaningful baseline, we integrate TransPose~\cite{TransPoseSIGGRAPH2021} with MotionGPT~\cite{jiang2023motiongpt} as a toy system for both motion capture and analysis via sparse inertial signals, taking SMPL motion representation~\cite{SMPL2015} as the intermediate. To demonstrate the capability of our method in precisely analyzing detailed and diverse human motion via sparse signals compared to vision-based approaches, we also compare our method with two well-known open-source Vision-Language Models (VLMs)~\cite{wang2024internvideo2,chen2024motionllm} on motion description and instruction tasks. As shown in Tab.~\ref{tab:analysis}, our method consistently outperforms the baseline method and InternVideo2~\cite{wang2024internvideo2} in both tasks and achieves performance comparable to MotionLLM~\cite{chen2024motionllm}. This demonstrates the advantage of our method in motion understanding and analysis via IMU signals, which are far sparser than vision modalities, significantly facilitating downstream real-time applications. 

\begin{figure}[t]
  \centering
  \includegraphics[width=\linewidth]{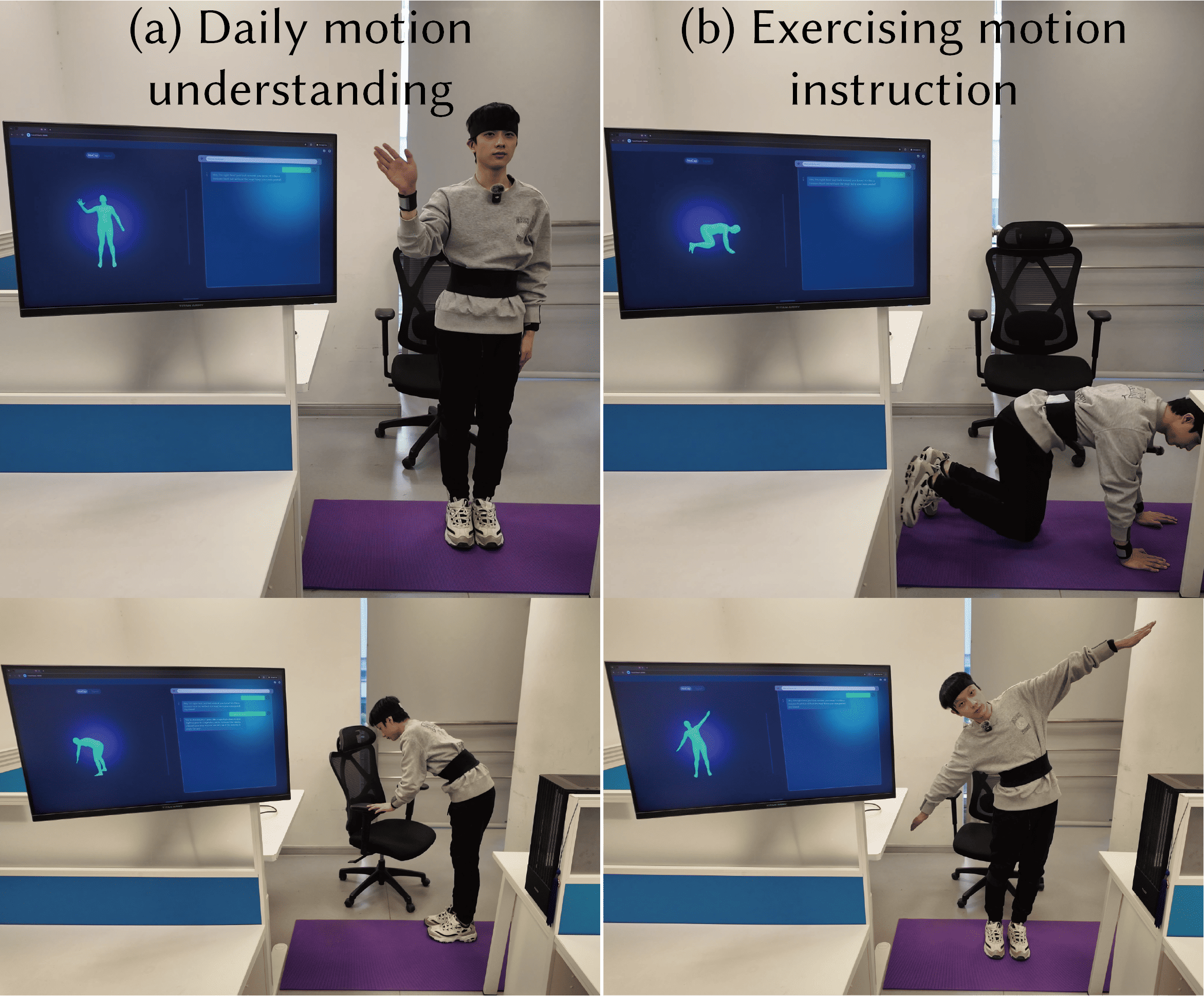}
  \captionof{figure}{\textbf{Web-based live demo.} 
  Our system processes streaming IMU signals in real time, enabling simultaneous motion capture and analysis. The live recording of a human performer in real world is displayed alongside the replicated virtual motions in the web interface, which also includes an integrated online chat window.}
\label{fig:real_time}
\end{figure}

\subsection{Web-based Live Demo}
In addition, we developed a web-based live demonstration for motion capture and analysis, powered by six IMU sensors and integrated with an LLM. As illustrated in Fig.~\ref{fig:real_time}, we present two primary application scenarios of our method: daily motion understanding and exercising motion instruction, providing both visual and textual feedback. The web demonstration follows the inference pipeline depicted in Fig.~\ref{fig:inference_pipeline}. Specifically, six Movella DOT IMU sensors~\cite{Movella} are attached to the corresponding body parts, and their connection is established via Bluetooth transmission. The user begins by performing a T-pose for approximately 10 seconds to calibrate the sensors. Once calibration is complete, the streaming signals are processed and fed into our IMU tokenizer, which generates jitter-reduced inertial tokens. Our method then employs the learned motion decoder to reconstruct the user's motion, with a post-processing module utilizing SmoothNet~\cite{zeng2022smoothnetplugandplaynetworkrefining}. Simultaneously, the IMU signals are continuously analyzed in the backend. For practical interaction, we monitor microphone input to capture the user's verbal questions and utilize Whisper-base-en~\cite{radford2023robust} to transcribe the audio into text. Finally, the textual feedback generated by our method is displayed in the chat window and converted into speech audio using SpeechT5-TTS~\cite{ao2021speecht5}.

\begin{figure}[t!]
  \centering
  \includegraphics[width=\linewidth]{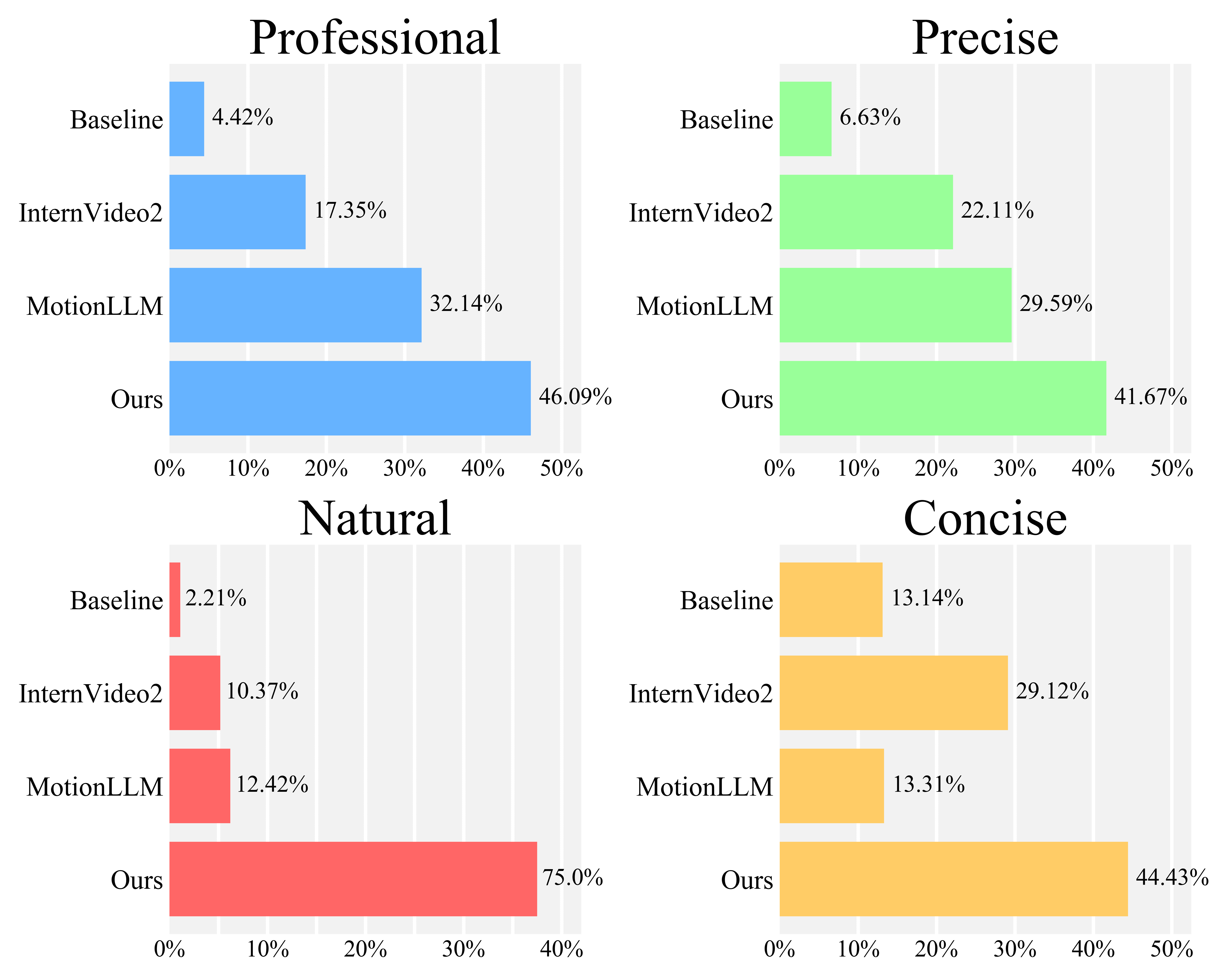}
  \captionof{figure}{\textbf{User study.} 
  A comparison of subjective preferences for textual feedback across four dimensions: Professionalism, Precision, Naturalism, and Brevity. Participants were asked to evaluate the textual feedback generated by four different methods (Baseline, InternVideo2, MotionLLM and Ours). Our approach achieved the highest ratings in all categories, demonstrating superior performance across all evaluated aspects.}
\label{fig:user_study}
\end{figure}

\subsection{User Study}

To comprehensively evaluate the performance and user experience of our system from a user perspective, we conducted a user study focusing on the professionalism, precision, naturalism, and brevity of system responses. The study compared four methods: the baseline, our method, MotionLLM~\cite{chen2024motionllm}, and InternVideo2~\cite{chen2024internvl2}. We distributed the study form to over 20 volunteers and invited participants to select their preferred options based on four different system responses, as well as the interaction processes demonstrated in pre-recorded videos. As illustrated in Fig.~\ref{fig:user_study}, our method secured approximately 50\% of the votes for providing accurate and reasonable feedback. Furthermore, in terms of naturalism, our method outperformed the other methods in user interactions.
\section{Discussions and Conclusions}
We have introduced Mojito, an innovative system for real-time human motion capture and online motion analysis, via jitter-reduced inertial tokens. By integrating a novel jitter-reduced and motion-aware IMU tokenizer with a large language model, Mojito establishes an interaction-friendly framework for motion description and instructor application. Experimental results demonstrate that our method achieves robust motion capture, effectively addressing various noisy input conditions that pose challenges for traditional RNN-based and physics solver-based approaches. Additionally, our user study highlights the professionalism, naturalism, precision and brevity of textual feedback generated by Mojito, underscoring its practical utility in real-world applications such as fitness training, rehabilitation and AR/VR.

\vspace{-4mm}
\paragraph{Limitations and Future Work}
Despite its contributions, Mojito still has several limitations. First, our jitter-reduced IMU tokenizer cannot operate in a per-frame inference manner, as it requires input signals to be segmented into data chunks. This leads to relatively high latency and discontinuous motion reconstruction results. Additionally, Mojito is constrained to structured IMU sensor placements on the human body, which may limit its applicability in more unstructured scenarios, such as general IMU signal understanding in robotics, autonomous driving, and object tracking.
In the future, we aim to explore autoregressive motion capture in real time through next-token prediction to mitigate inference latency and discontinuity. 
Furthermore, extend our multimodal system to general and unstructured IMU sensor signal understanding holds significant promise, as it could substantially broaden the scope of practical applications.

{\small
\bibliographystyle{ieee_fullname}
\bibliography{main}
}

\end{document}